# Proposal of Algorithms for Navigation and Obstacles Avoidance of Autonomous Mobile Robot

T. T. Hoang, D. T. Hiep, P. M. Duong, N. T. T. Van, B. G. Duong and T. Q. Vinh

Department of Electronics and Computer Engineering
University of Engineering and Technology
Vietnam National University, Hanoi
thuanhoang@donga.edu.vn

*Abstract*— **This paper presents algorithms to navigate and avoid obstacles for an in-door autonomous mobile robot. A laser range finder is used to obtain 3D images of the environment. A new algorithm, namely 3D-to-2D image pressure and barriers detection (IPaBD), is proposed to create a 2D global map from the 3D images. This map is basic to design the trajectory. A tracking controller is developed to control the robot to follow the trajectory. The obstacle avoidance is addressed with the use of sonar sensors. An improved vector field histogram (Improved-VFH) algorithm is presented with improvements to overcome some limitations of the original VFH. Experiments have been conducted and the result is encouraged.**

## I. INTRODUCTION

The navigation process for an autonomous mobile robot can be divided into four steps: environment mapping, trajectory design, motion control, and obstacle avoidance. In order to obtain good navigation performance, it is necessary to have two separated parts: the global and the local navigation. The global process uses information taken from sensors which have enough visibility of the environment such as the 3D laser and camera to generate a round path planning and control strategy. For non-structural environment like indoor, the robot must view the whole environment by itself to create a map and design a trajectory which can lead it to the destination. With the requirement of high precision, the image data from the camera is often insufficient due to its dependence on the light condition and the surface structures of objects. In addition, a vision system cannot directly measure the geometry parameters such distances between objects. The stereo camera can partly overcome this problem but the computation is expensive and the accuracy is low [1, 2]. Instead, the laser range finder is an appropriate choice. It measures distances at high rate and precision while insensitive to environment's conditions [3]. With the constraint that the robot always moves on the floor plane, the navigation task can be executed by directly using the pixel cloud of the 3D environment [4-5]. However, the algorithm is complicated and the computation is relatively high.

In this paper, we propose an algorithm which builds a 2D global map from 3D images for the navigation problem. The algorithm, namely 3D-to-2D image compression, based on the idea that pixels at a same position but different height can be presented by a single pixel described the occupancy or vacancy of the environment that the robot moves through. Based on the created 2D map, the A* algorithm is employed to design the global trajectory [6]. The motion control to follow the designed trajectory is derived by the Lyapunov stability criterion [7-9].

In the local navigation, the robot needs to react to the change of the environment such as a sudden appearance of obstacles. Sonar sensors have been proven to be effective for this task [10-13]. A well-known algorithm is *virtual force field method* [11, 13]. Nevertheless, this method has the week point that when the robot moves from a square to another in the net graph, the sum of propulsive forces is changed much in both the intensity and the direction so that there exists the case the robot cannot pass vacancy space such as an opened door because the propulsive forces from two sides of the door make a joint force which pushes the robot away from the door [12]. In order to overcome this problem, we applied the VFH method of J. Borenstein [13] and tuned it to be suitable to the robot's configuration. This method, called the Improved-VFH, uses the polar coordinate to control the robot to avoid obstacles in a region from 0,3m to 4m.

## II. CONSTRUCTING THE 2D MAP FROM THE CLOUD OF 3D PIXELS OF THE ENVIRONMENT

### A. Collecting data from laser sensor

Figure 1 shows the image of the laser range finder LMS-211 (LRF) in association with other sensors in a mobile robot designed at our laboratory. The measurement of the distance to a point in the obstacle is based on the principle of determining the going-returning time of the laser pulse reflected from the obstaclc. In order to get the 3D image of the environment, the LRF is stacked in a configuration that it can pitch up and down around a horizontal axis as described in [3]. In this work we adjust parameters of LRF so that every cycle of up pitching, the sensor gets 94 2D photo frames with pitch angles in the range from -5º to +20º and the increasing step of 0.266º. Each obtained 2D photo frame contains a data set corresponding to the horizontal scan of the laser ray with a scanning angle $\beta_k$ in the range from 40º to 140º and the resolution 1º. Consequently, the obtained 3D image is a cloud with 9,494 pixels. This image has the linearity in the height due to the constant rate of the pitching rotation.

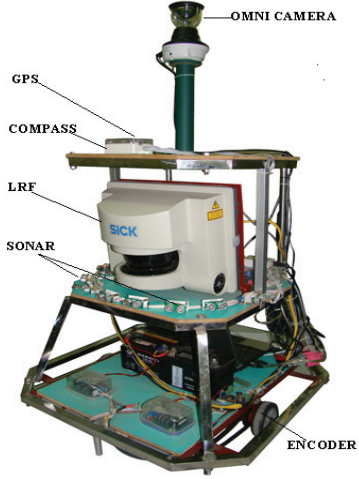

Figure 1. The laser range finder and sonar sensors in association with other sensors attached in a mobile robot.[14-15].

Since the sensor is put on the height 0.4m relative to the floor and at the minimum pitch angle -5º, the sensor only can detect objects at a distance at least 4.58m. Obstacles which are in the smaller distance to the robot are detected by eight sonar sensors. The combination of data from the laser range finder and sonar sensors gives the robot an image of the whole environment.

### B. Constructing the 2D global map from 3D laser images

At a pitch angle $\alpha_j$, the robot obtains a 2D image frame consisting of pixels determined by the horizontal angles $\beta_k$ and the distance $R_k$ to the object. From a set of 2D image frame, Cartesian coordinates of pixels that creats the 3D image of the environment are determined as below [3].

$$\begin{aligned} x_{j,k} &= R_k \cos\alpha_j \cos\beta_k \\ y_{j,k} &= R_k \cos\alpha_j \sin\beta_k \\ z_{j,k} &= R_k \sin\alpha_j \end{aligned} \quad (1)$$

The projection of pixels of objects on a plane x-y parallel with the floor can be used to construct the 2D map. Taking the union (U) of all pixels which have same coordinates (x, y) along the direction z of the environment creates a unique 2D map on the plane x-y as shown in figure 2.

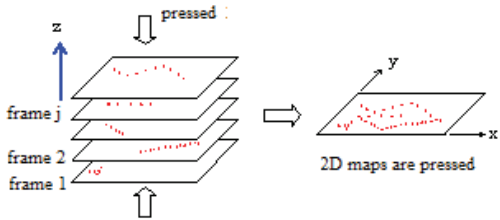

Figure 2. Pressure of frames in the 3D image to get the 2D map

It is a matrix of pixels in which every pixel is expressed by a pair of parameters $(\beta_k, R_k)$, where $\beta_k$ is the scanning angle of the k-th laser ray and $R_k$ is the distance measured to the pixel. The relation between these two parameters is not monovalent in each 2D frame. It means that there is possible one or many points with different $R_k$ (associated to different heights of z) in correspondence to a scanning horizontal angle $\beta_k$.

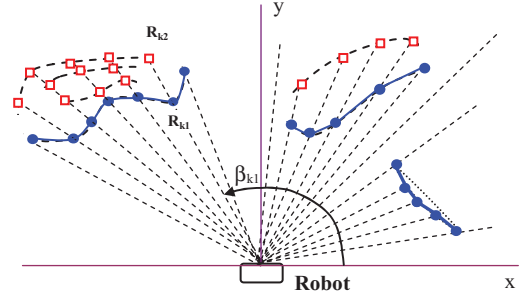

Figure 3. Detecting pixels which have the minimum distance (circular points in seamless lines)

The generated 2D image presents the surface image of obstacles in front of the robot. It is necessary to determine the boundary of image regions as they will be use for the path planning and obstacle avoidance. As shown in fig.3, this can be done by selecting pixels (circle points) that have the smallest distance to the robot among pixels received at the same laser scanning angle. The remaining pixels (squared points) are neglected. After this processing step, the relation between β and R becomes monovalent. The segmentation algorithm is then applied. The PDBS method is employed [16]. To further simplify the image, it is unnecessary to process the pixels at the height higher than the robot height as those pixels do not affect to the robot navigation. Let $z_{limit}$ be the robot height. We can remove all pixels with the height higher then $z_{limit}$. In summary, all steps to construct the 2D map from a 3D image are described as follows:

Step 1: Starting from the cloud of 3D pixels, the union of all pixels is taken along the vertical direction z.

Step 2: For each horizontal scanning angle $\beta_k$, find and select the value $R_k = R_{min}$.

Step 3: Remove points with $z > z_{limit}$ and z = 0 (corresponding to the floor).

Step 4: Performing the image segmentation algorithm (if necessary).

We call this algorithm 3D-to-2D image pressure and boundary detection (IPaBD).

### III. TRAJECTORY DESIGN AND MOTION CONTROL

#### A. Trajectory design

The next step is to find the optimal path for the robot to reach the destination from the initial position (the path planning).

In IPaBD algorithm, scanning pixels that present the surrounding environment of the robot can be described as a set of line segments as in [17] where line segments are obstacles such as wall, door, or lobby…Based on these line segments, the floor space is discretized into grid-cell map in which the size of each cell is set to be $(a \times a)$ cm$^2$, where $a$ equals to 1/3 diameter of the robot chassis. This grid map can be presented as a matrix with elements 0 and 1 where element 0 implies the appearance of obstacle. A "dilation" process is used in occupied region to extend two more cells to ensure that the robot cannot collide with the occupied region.

Each line segment extracted from scanning pixels is characterized in the Cartesian coordinate by two nodal points. Line segment parameters such as length, line equation, and angular coefficient can be easily determined from these two nodal points. In grid map, cells of line segment created by two nodal points have the value of 1. Let $(x_{1,L_i}, y_{1,L_i})$ and $(x_{2,L_i}, y_{2,L_i})$ be coordinates of two nodal points presenting the i$^{th}$ obstacle $(i=1...n)$, the coordinates of these two points in grid map are calculated as below:

$$x_{j,G_i} = \frac{x_{j,L_i} - x_{min}}{cellsize} \qquad y_{j,G_i} = \frac{y_{j,L_i} - y_{min}}{cellsize}$$
$$(i=1...n) \qquad (j=1...2)$$

where *cellsize* is the size of the grid cell. Cells occupied by the line segment presented by $x_{j,Gi}$ and $y_{j,Gi}$ are determined according to the following condition:

$$\left| \frac{m - x_{1,G_i}}{n - y_{1,G_i}} - \frac{x_{2,G_i} - x_{1,G_i}}{y_{2,G_i} - y_{1,G_i}} \right| \leq T \text{ if } x_{1,G_i} \neq x_{2,G_i}$$

$$m - x_{1,G_i} \leq T \text{ if } x_{1,G_i} = x_{2,G_i}$$

where *(m, n)* is the coordinate of cell in grid map and T is the size of the robot.

From data of occupied cells, we employed the A* algorithm [6] to find the shortest path to the destination D($x_D$, $y_D$) from the starting cell S($x_S$, $y_S$) (figure 9).

*B. Motion control*

After the path planning, control algorithm need developing to navigate the robot to follow the designed trajectory. In this paper we introduce a new control algorithm as follows.

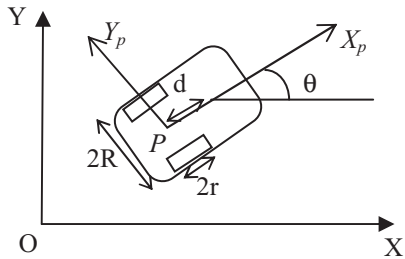

Figure 4. A noholonomic mobile robot platform.

Let the tangent and angular velocities of the robot be $v$ and $\omega$, respectively. We have:

$$\mathbf{v} = \begin{bmatrix} v_r \\ v_l \end{bmatrix} = \begin{bmatrix} \frac{1}{r} & \frac{R}{r} \\ \frac{1}{r} & \frac{-R}{r} \end{bmatrix} \begin{bmatrix} v \\ \omega \end{bmatrix}; \dot{\mathbf{q}} = \begin{bmatrix} \dot{x} \\ \dot{y} \\ \dot{\theta} \end{bmatrix} = \begin{bmatrix} \cos\theta & 0 \\ \sin\theta & 0 \\ 0 & 1 \end{bmatrix} \begin{bmatrix} v \\ \omega \end{bmatrix} \qquad (2)$$

The objective of the control problem is to design an adaptive control law so that the position vector $\mathbf{q}$ and the velocity vector $\dot{\mathbf{q}}$ track the desired position vector $\mathbf{q}_r(t)$ and velocity vector $\dot{\mathbf{q}}_r(t)$ under condition that robot parameters are not exactly known. The desired position and velocity vectors are represented by:

$$\mathbf{q}_r = [x_r \ y_r \ \theta_r]^T$$
$$\dot{x}_r = v_r \cos\theta_r$$
$$\dot{y}_r = v_r \sin\theta_r \qquad \text{with } v_r > 0 \text{ for all } t \qquad (3)$$
$$\dot{\theta}_r = \omega_r$$

The tracking error position between the reference robot and the actual robot can be expressed in the mobile basis frame as below [18]:

$$\mathbf{e}_p = \begin{bmatrix} e_1 \\ e_2 \\ e_3 \end{bmatrix} = \begin{bmatrix} \cos\theta & \sin\theta & 0 \\ -\sin\theta & \cos\theta & 0 \\ 0 & 0 & 1 \end{bmatrix} \begin{bmatrix} x_r - x \\ y_r - y \\ \theta_r - \theta \end{bmatrix} \qquad (4)$$

The derivation of the position tracking error can be expressed as:

$$\dot{\mathbf{e}}_p = \begin{bmatrix} \dot{e}_1 \\ \dot{e}_2 \\ \dot{e}_3 \end{bmatrix} = \begin{bmatrix} \omega e_2 - v + v_r \cos e_3 \\ -\omega e_1 + v_r \sin e_3 \\ \omega_r - \omega \end{bmatrix} \qquad (5)$$

There are some methods in the literature to select the smooth velocity input. In this research, we choose a new control law for $v, \omega$ as:

$$\begin{bmatrix} v \\ \omega \end{bmatrix} = \begin{bmatrix} v_r \cos e_3 + k_1 e_1 \\ \omega_r + k_3 e_3 + v_r e_2 \frac{\sin e_3}{e_3} \end{bmatrix} \qquad (6)$$

where $k_1, k_3 > 0$. One can see that when $e_3 \to 0$ then $\frac{\sin e_3}{e_3} \to 1$ so $\omega$ always be bounded.

With the control law in equation (6), it is easy to prove the asymptotical stability of the system. Choosing a positive definite function $V_p$ as follows:

$$V_p = \frac{1}{2} \mathbf{e}_p^T \mathbf{e}_p = \frac{1}{2}(e_1^2 + e_2^2 + e_3^2) \qquad (7)$$

The derivation of $V_p$ with respect to time $\dot{V}_p$ is:

$$\dot{V}_p = \mathbf{e}_p^T \dot{\mathbf{e}}_p = e_1\dot{e}_1 + e_2\dot{e}_2 + e_3\dot{e}_3$$
$$= e_1(\omega e_2 - v + v_r\cos e_3) + e_2(-\omega e_1 + v_r\sin e_3) + e_3(\omega_r - \omega) \quad (8)$$
$$= e_1(-v + v_r\cos e_3) + e_2 v_r\sin e_3 + e_3(\omega_r - \omega)$$

Replacing (6) into (8) gives
$$\dot{V}_p = e_1(-v + v_r\cos e_3) + e_2 v_r\sin e_3 + e_3(\omega_r - \omega)$$
$$= e_1\left(-v_r\cos e_3 - k_1 e_1 + v_r\cos e_3\right) + e_2 v_r\sin e_3 + e_3\left(\omega_r - \omega_r - k_3 e_3 - v_r e_2\frac{\sin e_3}{e_3}\right) \quad (9)$$
$$= -k_1 e_1^2 - k_3 e_3^2$$

It is able to see that $\dot{V}_p$ is continuous and bounded according to the Barbalat theorem. It means that $\dot{V}_p \to 0$ when $t \to \infty$. Consequently, $e_1 \to 0, e_3 \to 0$ when $t \to \infty$. Applying the Barbalat theorem again for the derivation, we get:
$$\dot{e}_1 \to 0, \dot{e}_3 \to 0 \quad (10)$$

and the equation (6) becomes
$$v \to v_r \quad (11)$$
$$\omega \to \omega_r \quad (12)$$

Combining (5), (10), (11), (12) gives: $e_2 \to 0, \dot{e}_2 \to 0$. Thus the control law (6) assures the proximity control system $\mathbf{e}_p \to 0$ when $t \to \infty$.

### C. Obstacle avoidance

As discussed in the section I, we developed a method for obstacle avoidance namely Improved-VFH. This method uses a net chart as in VFH but the updating law is changed as described in Figure 5. At each time reading information from the ultrasound sensor, the algorithm assigns values of 2 or 3 to squares as in the figure, other squares are assigned the value 0.

The data switch law from the net chart to the polar chart is inherited from VFH. After that, the process of calculation of the directional angle $\theta_d$ is divided into 2 steps as follows:

*Step 1*: Calculate slots with restriction in the range of sectors such that $|\theta - \theta_d| < 100^o$. If there exists a slot satisfying $|\theta - \theta_d| < 100^o$, the robot will control the directional angle so that $\theta = \theta_d$.

*Step 2*: If there is not any slot satisfying the above condition, the robot will perform a left or a right rotation with an angle $100^o$. The robot will rotate left if $\theta < \theta_t$ and vice verse, where $\theta_t = \text{atan2}(y_t - y, x_t - x)$ with $x_t, y_t, x, y$ are coordinates of the target and the current coordinates of the robot, respectively. The algorithm then search a new slot in the range $\theta \pm 100^o$, i.e. return to the Step 1.

If in the range $\theta \pm 100^o$ there exits more than one slot, then the algorithm will select a slot depending on each case:

Case 1: If $|\theta - \theta_t| < 90^o$, the algorithm selects a slot which has the directional angle $\theta_d$ so that $|\theta_d - \theta_t|$ is minimum.

Case 2: If $|\theta - \theta_t| \geq 90^o$, the algorithm selects a slot which has the directional angle $\theta_d$ so that $|\theta_d - \theta|$ is minimum.

The algorithm of control of the angle velocity is selected in a simple way as follows: $\omega = 10(\theta_d - \theta)$ and $|\omega| \leq 25^o/s$.

The algorithm to control the tangent velocity is selected as follows:
- If the robot is far from the obstacle, $V=V_{max}=0,5$m/s.
- If the robot is near the obstacle, $V=5(d_{30}-0.4)$ m/s, where $d_{30}$ is the distance from the robot to the nearest obstacle in the range $-30^o \to 30^o$ in the polar coordinate system attached to the robot.
- If $\omega \geq 10^o/s$, the tangent velocity V is replaced by $V^* = \dfrac{V}{2,5}$.

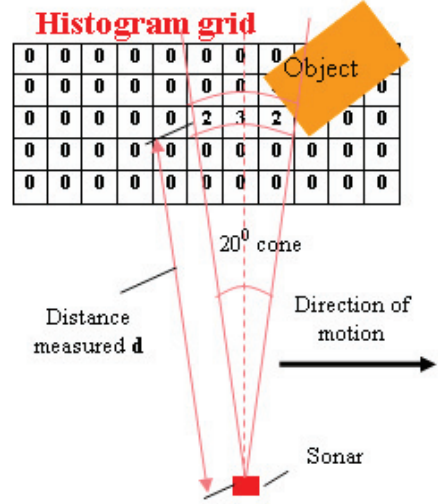

Figure 5. Environment information updating structure in the net chart in Improved VFH.

## IV. EXPERIMENT AND DISCUSSION

### A. The result of constructing the map by the IPaBD algorithm

Figure 6 shows an 3D image obtained by the laser range finder with the starting point S(0,0) and the destination D(0,7.6). In the image, there is a gate (A) with the girder shorter than the robot height and a gate (B) with the girder higher, where the robot height is 1.2m in the experiment. There is also a corridor next to the gates with the height that the robot can go through. The robot has to choose itself the shortest path from the starting position S to the destination position D without colliding obstacles.

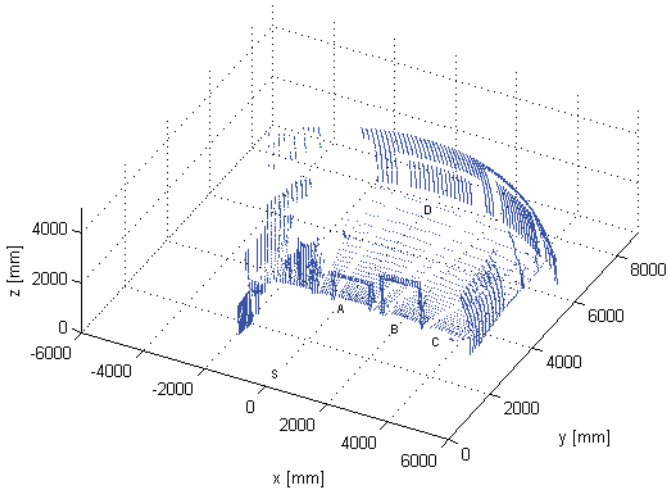

Figure 6. The global 3D image of the environment

Figure 7 shows the compressed 2D map obtained by applying the union (U) operator to 9494 pixels of the 3D image. Figure 8 presents the map constructed by the IPaBD algorithm.

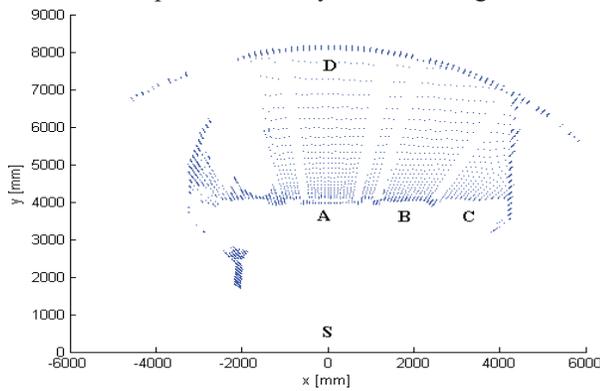

Figure 7. The 2D map with all 3D pixels pressed onto the plane x-y.

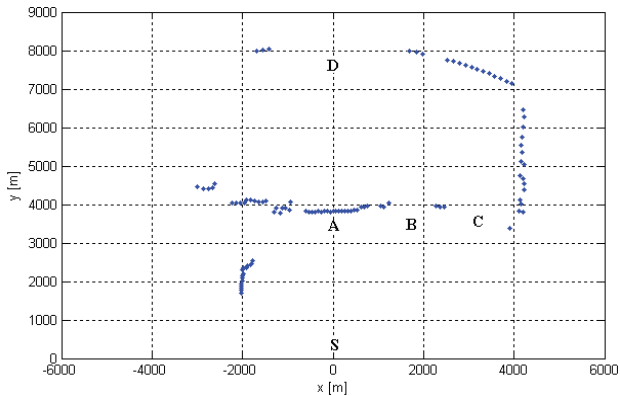

Figure 8. The result of the obtained map by the IPaBD algorithm.

### B. Path planning with A* algorithm

From the 2D map, the A* algorithm is applied to find the path from the starting position S to the destination D. As shown in figure 9, the map is divided into squares with the size of 40x40cm$^2$. Since the size of the robot is 80x80cm$^2$, it is necessary extend the occupied regions (presented by the grey cells) to avoid the collision. The result path shows that the robot does not pass the gate (A) since it detects an obstacle which is the girder. It also does not pass the corridor (C) as the path is long. Instead, it goes through gate (B) as expected.

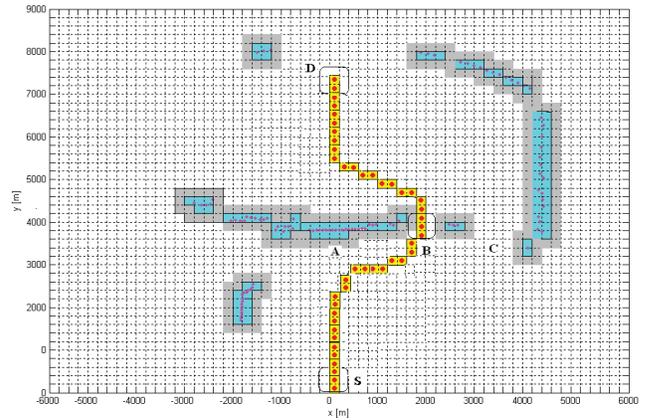

Figure 9. Robot navigation by the algorithm A*.

### C. Control the robot to track the trajectory

With the designed trajectory, we used the control law (6) to control the robot to follow the trajectory. The result is shown in figure 10. In the experiment, the tangent velocity of the robot is around 0.2m/s to 0.4m/s.

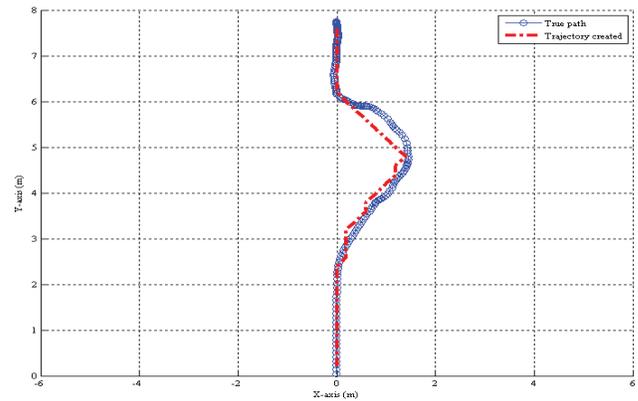

Figure 10. The real trajectory of the robot(the red curve) sticks to the designed trajectory(the blue curve).

### D. Avoiding obstacles using ultrasonic sensors

The experiment is implemented by sending an unexpected obstacle O to the designed trajectory. Figure 11 shows that the robot avoids the obstacle and then continues to track the global trajectory to arrive at the destination. Figure 12 is a sequence of images showing the robot motion during the experiments.

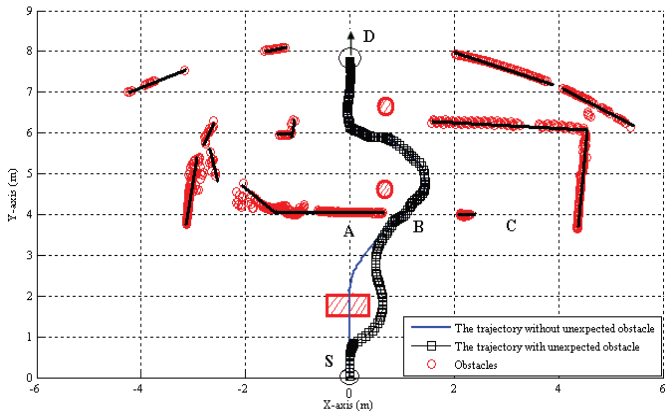

Figure 11. Avoiding an unexpected obstacle on the trajectory of the robot.

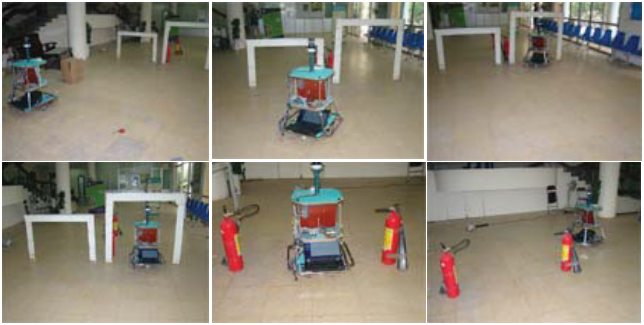

Figure 12. A sequence of images showing the robot motion during the experiments.

## V. CONCLUSION

This paper presents new algorithms for the problem of navigation and obstacle avoidance for the autonomous mobile robot. The navigation is developed from the environment data extraction to the mapping, path planning and trajectory tracking. The obstacle avoidance is accomplished with both known and unknown obstacles. The main contribution of the paper is the proposal of the IPaBD mapping algorithm for the path planning, the tracking control law for the robot system with partial unknown robot parameters, and the Improved-VHF algorithm for the obstacle avoidance. Experiments in a real robot system have been conducted and the result confirmed the effectiveness of the proposed methods.

ACKNOWLEDGEMENT

This work was supported by Vietnam National Foundation for Science and Technology Development (NAFOSTED).